\documentclass[runningheads]{llncs}

\usepackage{eccv}

\usepackage{eccvabbrv}

\usepackage{booktabs}
\usepackage{algorithm}
\usepackage{algpseudocode}
\usepackage[accsupp]{axessibility}  
\usepackage{multirow}

\usepackage{adjustbox} 

\usepackage{hyperref}

\usepackage{orcidlink}

\usepackage{graphicx}
\usepackage{tikz}
\usetikzlibrary{arrows.meta,calc,backgrounds}
\usepackage{xcolor}
\begin{document}

\title{Efficient Document Tampering Localization with Multi-Level Discrepancy Features and Unified DCT–Quantization Embedding} 

\titlerunning{Efficient Document Tampering Localization }

\author{Mohamed Dhouib\inst{1}\orcidlink{0000-0002-5587-1028} \and
Ye Zhu \inst{1}\orcidlink{0000-0003-0774-9375} \and
Sonia Vanier\inst{1}\orcidlink{0000-0001-6390-8882} \and
Aymen Shabou\inst{2}\orcidlink{0000-0001-8933-7053}}

\authorrunning{M.~Dhouib et al.}

\institute{LIX, École Polytechnique, IP Paris, France \and
DataLab Groupe, Crédit Agricole S.A., France\\
\email{mohamed.dhouib@polytechnique.edu}}

\maketitle
\begin{abstract}
Localizing document tampering is extremely challenging, as manipulations are crafted to appear visually consistent and often leave only subtle traces that are nearly invisible to the human eye. In prior work, evaluation has been largely dominated by synthetic benchmarks that closely match the training distribution, and methods have shown steady progress under this setting. However, these gains often translate poorly to human-made forgeries and to cross-domain evaluation, where both the source documents and the tampering pipeline can change, leading to a distribution shift. In addition, since the introduction of the Frequency Perception Head for the discrete cosine transform (DCT) modality, it has become a standard choice, and subsequent work has largely focused on downstream modules and fusion strategies rather than revisiting the backbone itself. To help close this gap in cross-domain performance and improve the DCT backbone design, we propose \textbf{DiffNet}, a relatively simple yet effective RGB--DCT early-fusion architecture driven by two key design choices. First, to ensure that the decoder aggregates multi-scale inconsistency evidence rather than operating on raw, content-heavy activations, we apply a lightweight multi-level discrepancy transformation at the output of each backbone stage, replacing features with magnitude-only responses to learned zero-sum filters. Second, we design an efficient DCT-domain backbone that relies on a lightweight frequency-index-aware DCT--quantization joint embedding. Our approach achieves state-of-the-art performance on cross-domain and human-made document tampering localization, outperforming prior methods by around 30\%, with up to $7\times$ higher throughput than the previous best model.

\end{abstract}
\section{Introduction}
Document images are central to financial, administrative, and identity-related workflows, making them attractive targets for tampering. Such manipulations, if undetected, can enable fraud and identity theft, undermine legal evidence, and trigger costly downstream errors in automated decision-making systems. Yet state-of-the-art models~\cite{dtdv2freq,rtm,adcd-net} still generalize poorly to cross-domain and human-made forgeries, and evaluation has mainly been conducted on synthetic tampering benchmarks~\cite{Doctamper,adcd-net,dtdv2freq,neurips} that often remain close to the training distribution.

Because annotated human-made tampering data are scarce and costly to collect, prior work typically resorts to large-scale synthetic pretraining as a practical necessity. However, this setting encourages shortcut learning. Synthetic generation pipelines can introduce systematic signatures. Inserted text is generally rendered from a limited font set or a fixed generator family, while coverage often reuses a small set of algorithms that leave characteristic boundary or texture artifacts. Similar repetitive signatures arise in copy-move, splicing, and blending operations, enabling models to rely on pipeline-specific fingerprints instead of the underlying \emph{inconsistencies} that manipulations create and that are more likely to transfer across domains and forgery tools.

Moreover, synthetic datasets frequently come from a narrow set of source domains and acquisition pipelines, and for underrepresented document types such as payslips, bank statements, and tax forms, they often generate many manipulated variants from the same source image. For instance, \cite{leveragingcontrastive} draws a substantial portion of its source documents from IIT-CDIP~\cite{IITCDIP}, a collection originating from tobacco litigation materials which represents a highly specific provenance. As a result, models can reduce training loss in hard cases by exploiting domain-specific cues rather than focusing on the inconsistencies the tampering introduces. These issues can persist even in human-made document tampering datasets, which are often small~\cite{findit,finditagain} and produced by few experts.

This motivates us to tackle the task from a different direction: rather than further increasing architectural complexity, we constrain the feature pyramid so the decoder becomes mainly driven by inconsistency cues. Specifically, we introduce a lightweight \emph{multi-level discrepancy transformation} applied at the output of every backbone stage before multi-level fusion and prediction. It converts stage features into magnitude-only responses of learned zero-sum filters, yielding a sign-invariant discrepancy representation across the entire feature pyramid. As a result, multi-level fusion aggregates inconsistency cues across stages rather than operating on raw content-heavy activations.

Beyond RGB cues, recent work has emphasized frequency evidence, in particular DCT coefficients, and has developed several RGB--DCT fusion architectures~\cite{Doctamper,CAT-Netv2,rtm,dtdv2freq,adcd-net}. The Frequency Perception Head \cite{Doctamper} (FPH) has emerged as a standard choice in several recent state-of-the-art document tampering architectures \cite{dtdv2freq,adcd-net,Omni-IML}, which have mainly emphasized stronger fusion strategies and auxiliary enhancement modules, while leaving the encoder backbone mostly unchanged. In contrast, we propose a more efficient alternative to the Frequency Perception Head. Since both quantized DCT coefficients and quantization-table entries are discrete integers, we leverage the expressiveness of embedding layers to map each coefficient–quantization pair to a compact joint embedding, explicitly injecting the frequency index while avoiding early high-dimensional feature expansions and auxiliary blocks.

Together, these two design choices yield \textbf{DiffNet}, a simple and efficient architecture that improves both cross-domain accuracy and throughput. On cross-domain evaluation, our architecture improves upon the previous best by around 30\% while increasing throughput by up to $7\times$. Our contributions are as follows:

\begin{itemize}
\item We introduce a multi-level discrepancy transformation applied at every backbone stage output, converting features into magnitude-only responses of learned zero-sum filters and yielding a sign-invariant discrepancy representation across the feature pyramid.
\item We design an efficient DCT-domain backbone based on a frequency-index-aware joint embedding of quantized DCT coefficients and quantization tables, providing a strong and lightweight frequency branch for early fusion.
\item We run extensive ablations on both components and their main design choices, confirming the contribution of each part and supporting the design decisions within each component.
\item Combining these components in a simple RGB--DCT early-fusion architecture yields a new state-of-the-art on human-made document tampering detection and localization, improving over the previous best by around 30\% while substantially increasing throughput. The code and pretrained weights are available at \url{https://github.com/Mohamed-Dhouib/DiffNet}.

\end{itemize}
\section{Related Work}
\subsection{Document tampering detection and localization datasets}
Progress in document tampering localization has long been constrained by the limited availability of public, annotated data. Early works~\cite{old0,old1,old2} were therefore developed and evaluated on private collections. \cite{wang2022b} introduced T-SROIE, a small benchmark centered on character-level replacements produced by SRNet~\cite{SRNET}. While useful for evaluation, such synthetic edits can exhibit systematic visual artifacts and may overestimate real-world performance. Later human-made datasets such as FindIt~\cite{findit}, FindItAgain~\cite{finditagain}, and RTM~\cite{rtm} were introduced, providing human-made manipulations that better reflect real-world scenarios, but remain comparatively limited in scale relative to synthetic corpora, making them more suitable for evaluation and fine-tuning than for pretraining. To enable large-scale training, DocTamper~\cite{Doctamper} constructed a substantially larger dataset via a rule-based generation pipeline. The authors also introduce two test splits intended for cross-domain evaluation. However, the source documents visual characteristics remain similar to those used to generate the training set, and the tampering is still produced by the same synthesis pipeline. Recently, \cite{leveragingcontrastive} proposed a learning-guided generation pipeline that trains auxiliary networks to produce higher-quality and more diverse tampered documents. Although the quality has improved compared to previous synthetic datasets, the pipeline used still introduces pipeline-specific traces that models can use as tampering cues, rather than relying on the inconsistencies they introduce. Nonetheless, large-scale pretraining on synthetic data is necessary given the scarcity of human-made forgeries. We therefore propose to constrain the feature pyramid, reducing its reliance on pipeline-specific artifacts.

\subsection{Document tampering detection and localization architectures}
DCT-domain frequency evidence has proven particularly effective for exposing manipulation traces in document images. DTD~\cite{Doctamper} introduced a dual RGB and DCT architecture whose Frequency Perception Head (FPH) first processes the quantized DCT coefficients before injecting quantization-table information. Concretely, each discrete coefficient is passed through a dilated convolution that expands the signal to a high-dimensional feature map of 64 channels at the initial image resolution. The result is then projected and modulated by an embedded quantization table. Next, the modulated and unmodulated features are concatenated, augmented with explicit coordinate channels, and fed to a strided convolution that expands the representation to 256 channels, followed by several MBConv blocks \cite{efficientnet}. This sequence materializes large intermediate activations early, contributing significant overhead in the DCT branch. FPH was reused by state-of-the-art models such as FFDN~\cite{dtdv2freq} and ADCD-Net~\cite{adcd-net}. FFDN~\cite{dtdv2freq} further strengthens spatial--frequency modeling and trace sensitivity with additional enhancement and fusion components. Other approaches further add specialized components~\cite{newdataandopenopen,dtdv3,dtdv4}, often at the cost of increased architectural complexity. More recently, ADCD-Net~\cite{adcd-net} improves robustness under practical post-processing that can disrupt JPEG block alignment. Overall, while these methods steadily improve performance, their initial evaluations and ablations of proposed components are often conducted on synthetic documents generated with the same pipeline as the training set, and their benefits have yet to be demonstrated for realistic human-made tampering and cross-domain settings, where both the source images and the manipulation pipeline can differ drastically. We therefore start from a standard early-fusion architecture and introduce a small set of well-motivated architectural changes aimed at improving generalization and providing a more efficient DCT backbone.

\subsection{Residual features}
In addition to leveraging frequency features, another common strategy in image forensics is to introduce residual-like operators as an additional modality alongside RGB. Hand-crafted residual features such as SRM~\cite{srm} and Bayar convolution layers~\cite{bayar2016,bayar2018} are used to produce high-pass representations that are fused with semantic features early on or processed in a parallel branch to improve sensitivity to subtle manipulation cues. While such auxiliary modalities can improve within-domain performance, recent unified evaluations suggest that they do not improve generalization~\cite{neurips}. In our work, we take a different direction and instead constrain the feature pyramid, rather than adding another modality.

\section{Method}
\begin{figure*}[t]
  \begingroup
  \def\INPAPER{1}
  \resizebox{\textwidth}{!}{\input{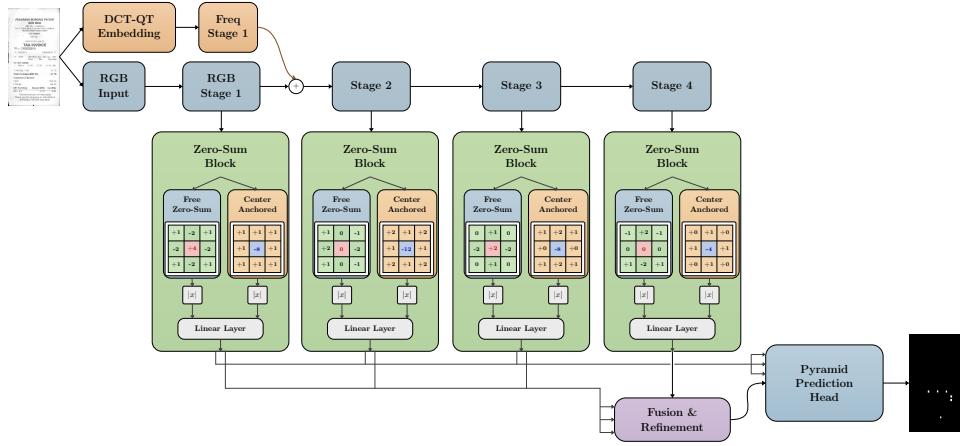}}
  \endgroup
  \caption{Overview of the proposed \textbf{DiffNet} architecture.}
  \label{fig:arch}
\end{figure*}
\subsection{Overview}
Our model follows a two-branch early-fusion design. The RGB branch processes the input image with a four-stage ConvNeXt-V2~\cite{convnext} backbone, producing a multi-scale feature pyramid. In parallel, the DCT branch encodes the quantized DCT coefficients together with the quantization table to produce a frequency feature map. We fuse the DCT features into the RGB stream after stage~1. At the output of each backbone stage, we apply our \emph{multi-level discrepancy transformation}, yielding a discrepancy feature pyramid, which is then aggregated by a multi-scale fusion module to predict the tampering mask. An overview of the architecture is shown in \cref{fig:arch}.

\subsection{Backbone Design}
\paragraph{RGB branch.}
Given an image $I \in \mathbb{R}^{H \times W \times 3}$, the RGB branch uses a ConvNeXt-V2 backbone to compute four stages of features $\{F_1, F_2, F_3, F_4\}$, with
$F_s \in \mathbb{R}^{C_s \times \frac{H}{2^{s+1}} \times \frac{W}{2^{s+1}}}$ for $s \in \{1,2,3,4\}$.
\paragraph{DCT branch.}
DCT coefficients together with quantization tables, which capture block-level frequency content and the per-frequency quantization strength, have proven to be an effective modality for document tampering detection~\cite{dtdv2freq,adcd-net}. Quantization and rounding introduce characteristic artifacts, often termed \emph{codec traces}, that have been shown to be important cues for tampering detection and are most naturally described by the pair of quantized DCT coefficients and the associated quantization table~\cite{Farid,pevny2008detection,dct,CAT-Netv2}. We therefore operate in the \emph{quantized} DCT domain. DCT maps each non-overlapping $8\times 8$ image block from the pixel domain to 64 frequency coefficients
in a cosine basis. Concretely, for a block $B\in\mathbb{R}^{8\times 8}$ with samples $B_{x,y}$, the unquantized 2D DCT coefficient at frequency
index $(u,v)\in\{0,\ldots,7\}^2$ is:
\begin{equation}
C_{u,v}=\alpha(u)\alpha(v)\sum_{x=0}^{7}\sum_{y=0}^{7}
B_{x,y}\cos\!\left(\frac{(2x+1)u\pi}{16}\right)\cos\!\left(\frac{(2y+1)v\pi}{16}\right),
\end{equation}
where $\alpha(0)=\frac{1}{\sqrt{8}}$ and $\alpha(t)=\sqrt{\frac{2}{8}}$ for $t\ge 1$. For each $8\times 8$ block and frequency index $(u,v)$, the raw quantized DCT coefficient is computed as:
\begin{equation}
\tilde{C}_{u,v} = \mathrm{round}\left(\frac{C_{u,v}}{Q_{u,v}}\right),
\end{equation}
where $C_{u,v}$ is the unquantized coefficient and $Q_{u,v}$ is the corresponding quantization table
entry. Let $\hat{C} \in \{0,\ldots,20\}^{\frac{H}{8}\times \frac{W}{8}\times 64}$
denote the stacked clipped absolute quantized DCT coefficients over all
$8 \times 8$ blocks, obtained by taking the absolute value of each raw
quantized DCT coefficient $\tilde{C}_{p,k}$ and clipping it to $[0,20]$, and let
$Q \in \{1,\dots,255\}^{64}$ denote the quantization table. We index blocks by $p$ and frequencies by
$k\in\{1,\dots,64\}$, and denote the quantization entry at frequency $k$ as $Q_k$. Since both $\hat{C}$ and $Q$ take discrete integer values, we use a learnable \emph{frequency-index-aware joint embedding} that maps each pair $(\hat{C}_{p,k}, Q_k)$ to a vector in $\mathbb{R}^{d_{\text{dct}}}$, with a small embedding size $d_{\text{dct}}$. Concretely, we embed the local quantized coefficient value into
$v_{p,k}\in\mathbb{R}^{d_{\text{dct}}}$ and we embed the corresponding quantization entry $Q_k$ and transform it into per-frequency modulation parameters
$\gamma_k,\beta_k\in\mathbb{R}^{d_{\text{dct}}}$, which modulate $v_{p,k}$ in a FiLM-like manner~\cite{FILM}. We additionally inject two complementary signals: (i) a learned frequency-index embedding $f_k\in\mathbb{R}^{d_{\text{dct}}}$ that identifies which DCT position $k$ the token comes from, and (ii) a global table bias $t\in\mathbb{R}^{d_{\text{dct}}}$ computed from the full quantization table $Q$, using an MLP, to capture quantization patterns beyond the single entry $Q_k$. The resulting per-frequency joint embedding is:
\begin{equation}
e_{p,k} = (1+\gamma_k)\odot v_{p,k} + \beta_k + f_k + t,
\end{equation}
where $\odot$ denotes element-wise multiplication. Stacking these embeddings for all frequencies yields a per-block tensor of shape $64\times d_{\text{dct}}$. We then obtain a single block-level representation by a simple reshape that concatenates the 64 frequency embeddings along the channel dimension, producing a vector in $\mathbb{R}^{64\,d_{\text{dct}}}$ per block. This produces an embedded tensor
\(
E \in \mathbb{R}^{\frac{H}{8}\times \frac{W}{8}\times C_{\text{dct}}}\) where \( C_{\text{dct}} = 64\,d_{\text{dct}}.
\)
We then process $E$ with a stack of standard ConvNeXt-V2 blocks~\cite{convnext}, producing a compact frequency feature map
\(
F_{\text{dct}} \in \mathbb{R}^{C_{\text{dct}} \times \frac{H}{8} \times \frac{W}{8}}
\).

\paragraph{Early fusion.}
We inject the DCT features into the RGB stream after the first stage by concatenation followed by a lightweight projection back to the RGB channel dimension. The remaining stages operate on the fused representation.
\subsection{Multi-level discrepancy transformation}
As discussed in the introduction, synthetic document tampering datasets used for pretraining can encourage models to treat domain-specific cues and pipeline-specific artifacts as evidence of tampering, rather than focusing on the \emph{inconsistencies} introduced by the manipulation. To reduce reliance on such cues and instead emphasize inconsistency evidence, we introduce a lightweight \emph{multi-level discrepancy transformation} applied at the output of every backbone stage before multi-level fusion and prediction. It converts stage features into magnitude-only responses of learned zero-sum filters, yielding a sign-invariant discrepancy representation across the feature pyramid.

\paragraph{Channel-wise zero-sum discrepancy filters.}
Let $F \in \mathbb{R}^{C \times H_i \times W_i}$ be a feature map at stage $i$. For each channel $c$ independently, we produce $M$ responses using $K\times K$ depthwise filters:
\begin{equation}
u_{c,m}(p) = \sum_{\Delta \in \{-\frac{K-1}{2},\ldots,\frac{K-1}{2}\}^2} k_{c,m}(\Delta)\, F_c(p+\Delta),
\qquad m=1,\ldots,M.
\end{equation}

\noindent We use zero-sum filters, i.e., $\sum_{\Delta} k_{c,m}(\Delta)=0$, which suppresses channel-wise bias components and emphasizes discrepancy responses rather than absolute feature intensity. We then take the absolute response, $v_{c,m}(p)=|u_{c,m}(p)|$, so the representation captures the \emph{strength} of the mismatch independently of its sign. We can split the kernel into positive and negative parts $k=k^+-k^-$ with
$k^+(\Delta)=\max(k(\Delta),0)$ and $k^-(\Delta)=\max(-k(\Delta),0)$.
Let $\mathcal{I}:=\{-\frac{K-1}{2},\ldots,\frac{K-1}{2}\}^2$ denote the set of spatial offsets.
Then we have:
\begin{equation}
\begin{aligned}
v_{c,m}(p)
&=\Bigl|\sum_{\Delta \in \mathcal{I}} k^+_{c,m}(\Delta)\,F_c(p+\Delta)
-\sum_{\Delta \in \mathcal{I}} k^-_{c,m}(\Delta)\,F_c(p+\Delta)\Bigr|,\\[-0.4em]
&\hspace{1.8em}\text{with }\sum_{\Delta \in \mathcal{I}} k^+_{c,m}(\Delta)
=\sum_{\Delta \in \mathcal{I}} k^-_{c,m}(\Delta).
\end{aligned}
\end{equation}

\noindent Since both terms have the same total weight, they form two competing weighted sums, and $v_{c,m}(p)$ captures their difference in magnitude, producing a discrepancy-strength signal.

\paragraph{Free and center-anchored zero-sum filters.}
We let the model learn two complementary types of zero-sum filters:
\begin{itemize}
\item \textbf{Free zero-sum filters} keep flexibility while enforcing $\sum_{\Delta} k_{c,m}(\Delta)=0$.
\item \textbf{Center-anchored zero-sum filters} further constrain the filter to behave as a neighborhood-based center predictor: neighbor weights are non-negative and the center weight is the negative sum of neighbors, so the output measures a center-to-context discrepancy.
\end{itemize}
These two families are complementary: the center-anchored family injects a structured inductive bias, while the free family preserves expressivity.

\paragraph{Unified parameterization.}
\label{sec:method}
We propose a single formulation that covers both families by tying the neighbor weights to the center weight, which will allow us to define an efficient implementation at training time.
Let $\mathcal{N}=\left\{-\frac{K-1}{2},\ldots,\frac{K-1}{2}\right\}^2 \setminus \{(0,0)\}$ denote the set of $K\times K$ offsets excluding the center. For each $(c,m)$ we learn unconstrained parameters $\theta_{c,m,\Delta}$ and define:
\begin{equation}
a_{c,m,\Delta} =
\begin{cases}
\theta_{c,m,\Delta} & \text{(free family)},\\
\lvert\theta_{c,m,\Delta}\rvert & \text{(center-anchored family)},
\end{cases}
\qquad \Delta \in \mathcal{N}.
\end{equation}

\noindent The convolution kernel can be formulated as follows:
\begin{equation}
k_{c,m}(\Delta) =
\begin{cases}
a_{c,m,\Delta} & \Delta \in \mathcal{N},\\
-\sum_{\delta \in \mathcal{N}} a_{c,m,\delta} & \Delta=(0,0).
\end{cases}
\end{equation}

\noindent which automatically enforces the zero-sum constraint and correctly defines both filter families. A direct implementation would explicitly synthesize the full $K\times K$ depthwise kernels inside the forward pass, introducing additional tensor construction overhead before the convolution. Our proposed parametrization allows us to define an efficient custom CUDA kernel, which we explain in~\cref{app:custom-op}. At inference time, once parameters are fixed, we materialize the resulting depthwise kernels once and use a standard depthwise convolution.

\paragraph{Projection.}
The $M$ magnitude responses per channel are then mapped back to the backbone channel dimension with a lightweight channel-wise projection, yielding the transformed feature map $\tilde{F}=\phi(F)$.

\paragraph{Multi-stage application.}
We apply the same transformation at the output of every backbone stage before multi-level fusion and prediction. For each stage $s$, we first project $F_s$ to a relatively small intermediate width and then apply $\phi$ in this space, producing the transformed feature map:
\begin{equation}
\tilde{F}_s = \phi\!\left(r(F_s)\right), \qquad s \in \{1,2,3,4\},
\end{equation}
where $r(\cdot)$ is a lightweight projection to a reduced channel dimension. The decoder and fusion modules then operate on $\{\tilde{F}_1,\tilde{F}_2,\tilde{F}_3,\tilde{F}_4\}$.

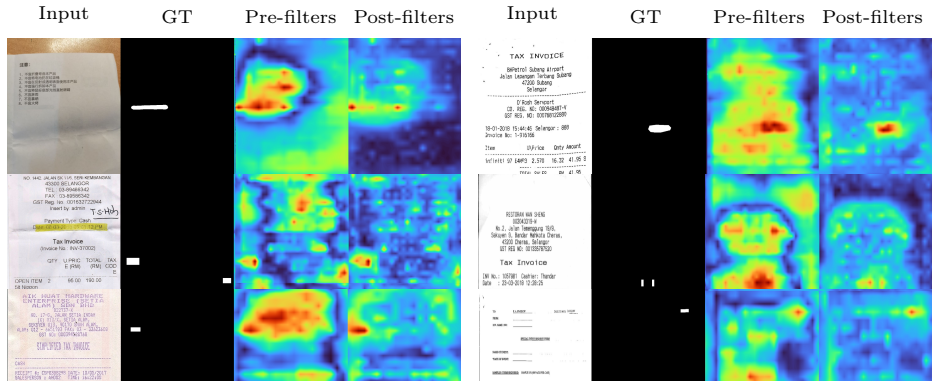
\begin{figure*}[t]
\centering

\begin{tikzpicture}
  \node[anchor=south west, inner sep=0] (tmp) {\rule{0.49\textwidth}{0pt}};
  \node[anchor=south, font=\scriptsize] at ($(tmp.north west)!0.125!(tmp.north east)+(0,1.2mm)$) {Input};
  \node[anchor=south, font=\scriptsize] at ($(tmp.north west)!0.375!(tmp.north east)+(0,1.2mm)$) {GT};
  \node[anchor=south, font=\scriptsize] at ($(tmp.north west)!0.625!(tmp.north east)+(0,1.2mm)$) {Pre-filters};
  \node[anchor=south, font=\scriptsize] at ($(tmp.north west)!0.875!(tmp.north east)+(0,1.2mm)$) {Post-filters};
\end{tikzpicture}\hfill
\begin{tikzpicture}
  \node[anchor=south west, inner sep=0] (tmp) {\rule{0.49\textwidth}{0pt}};
  \node[anchor=south, font=\scriptsize] at ($(tmp.north west)!0.125!(tmp.north east)+(0,1.2mm)$) {Input};
  \node[anchor=south, font=\scriptsize] at ($(tmp.north west)!0.375!(tmp.north east)+(0,1.2mm)$) {GT};
  \node[anchor=south, font=\scriptsize] at ($(tmp.north west)!0.625!(tmp.north east)+(0,1.2mm)$) {Pre-filters};
  \node[anchor=south, font=\scriptsize] at ($(tmp.north west)!0.875!(tmp.north east)+(0,1.2mm)$) {Post-filters};
\end{tikzpicture}

\vspace{-1.0mm}

\begin{tikzpicture}
  \node[anchor=south west, inner sep=0] (img) {\includegraphics[width=0.49\textwidth]{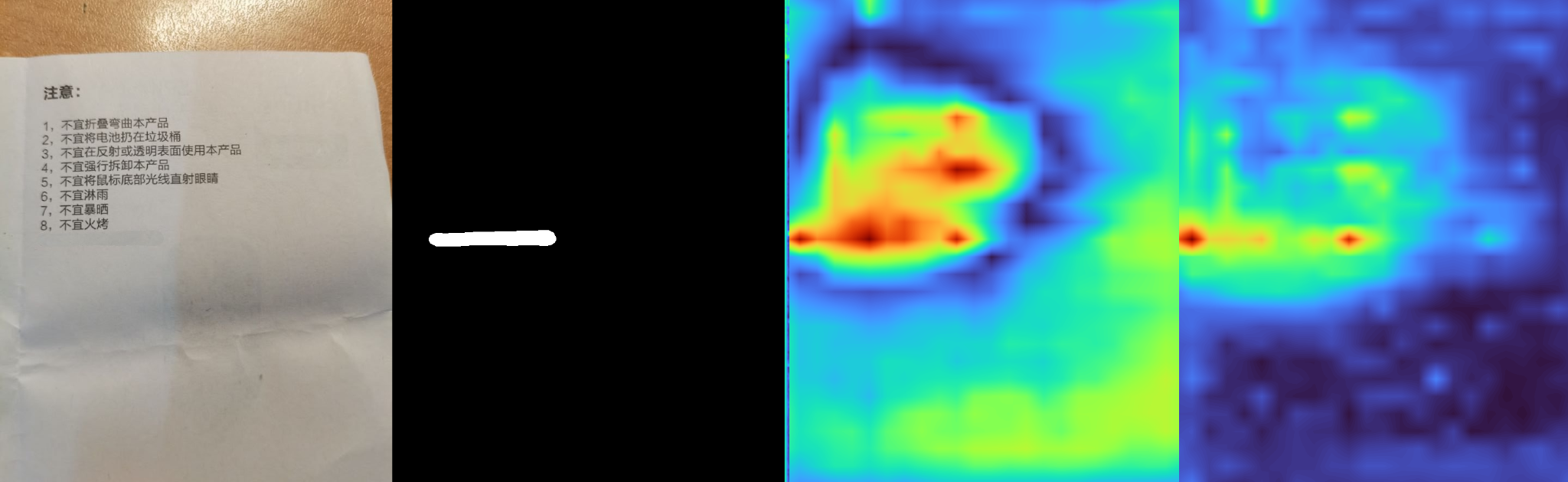}};
\end{tikzpicture}\hfill
\begin{tikzpicture}
  \node[anchor=south west, inner sep=0] (img) {\includegraphics[width=0.49\textwidth]{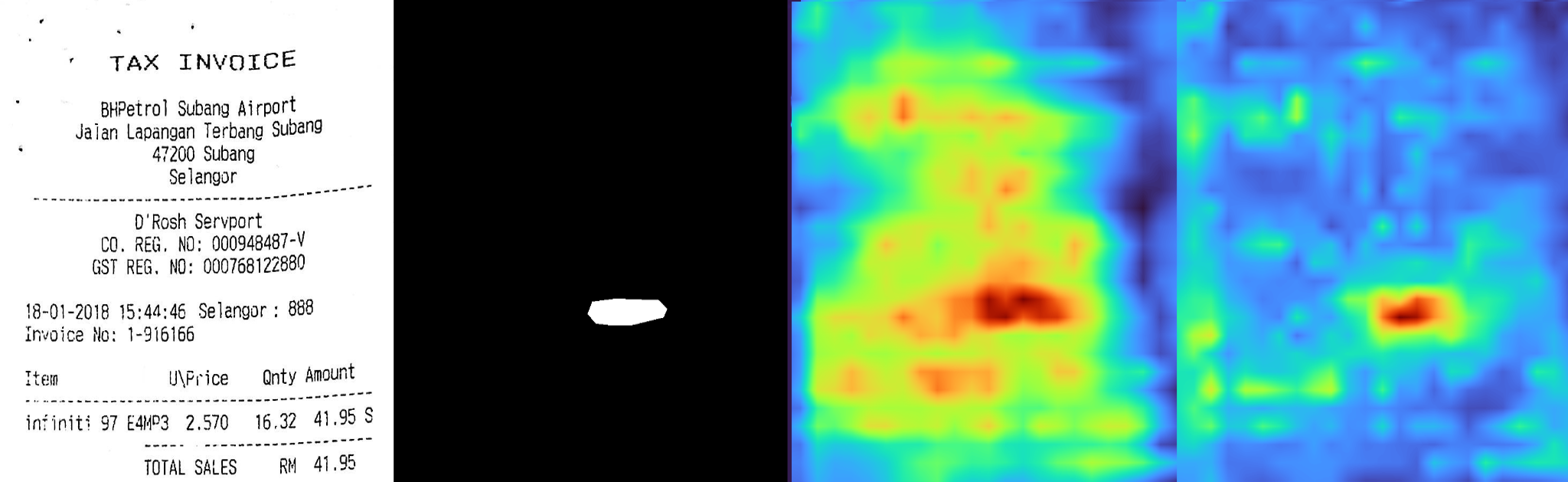}};
\end{tikzpicture}\\[-2pt]
\begin{tikzpicture}
  \node[anchor=south west, inner sep=0] (img) {\includegraphics[width=0.49\textwidth]{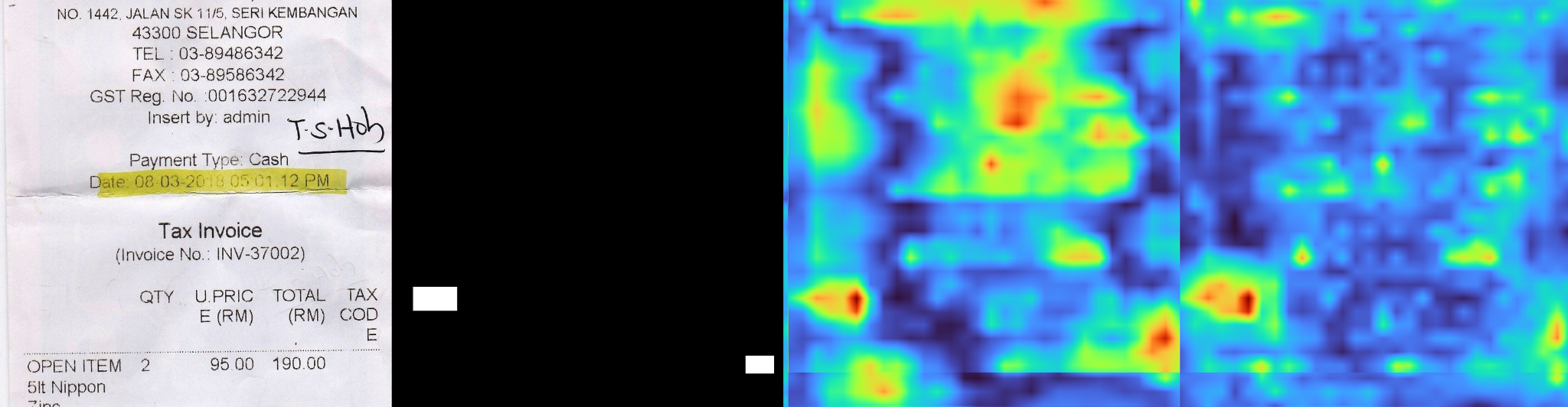}};
\end{tikzpicture}\hfill
\begin{tikzpicture}
  \node[anchor=south west, inner sep=0] (img) {\includegraphics[width=0.49\textwidth]{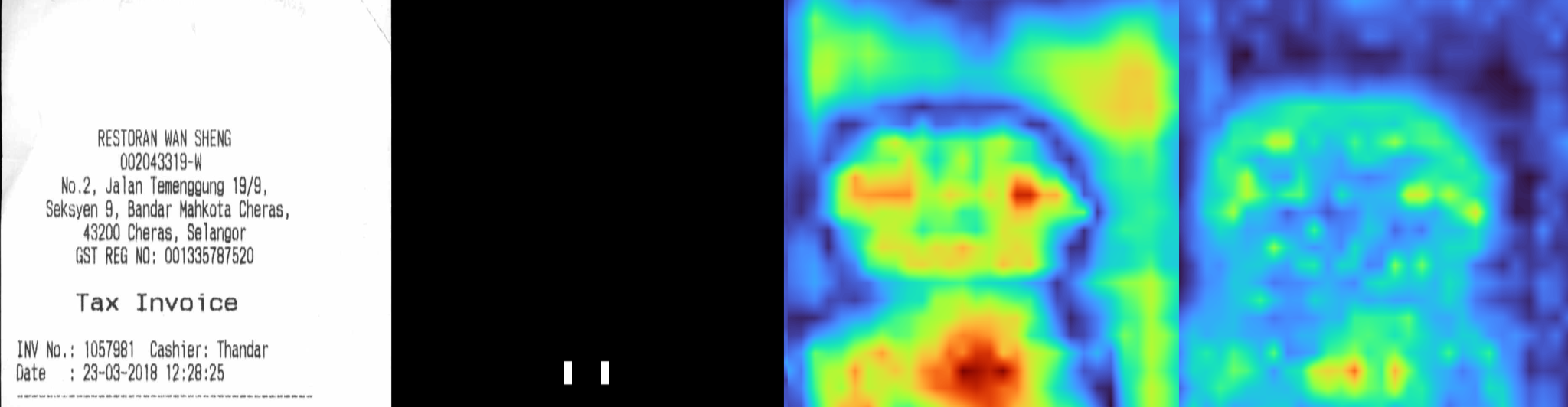}};
\end{tikzpicture}\\[-2pt]
\begin{tikzpicture}
  \node[anchor=south west, inner sep=0] (img) {\includegraphics[width=0.49\textwidth]{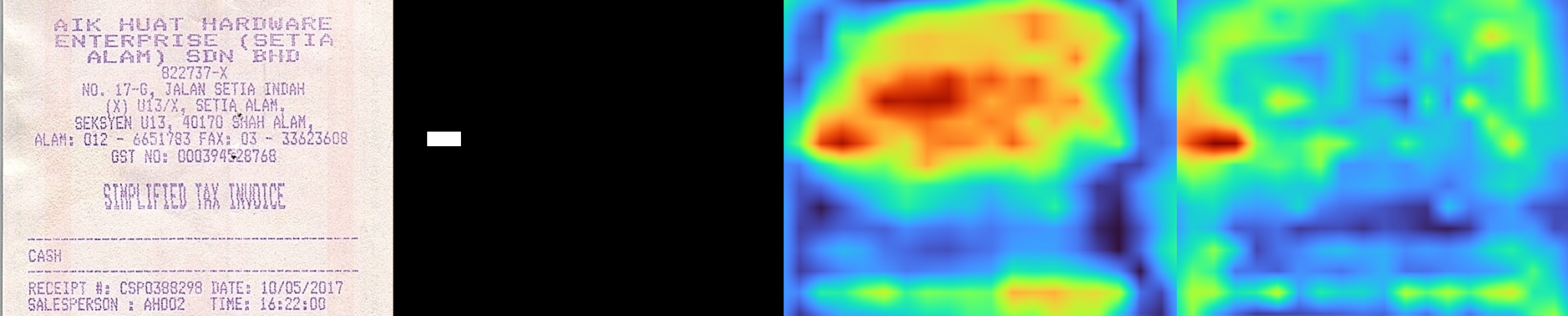}};
\end{tikzpicture}\hfill
\begin{tikzpicture}
  \node[anchor=south west, inner sep=0] (img) {\includegraphics[width=0.49\textwidth]{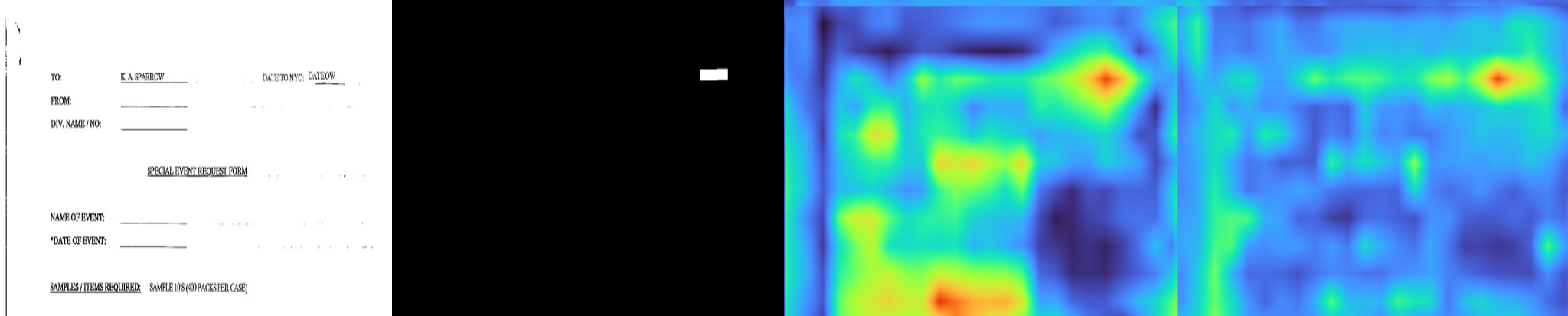}};
\end{tikzpicture}

\caption{Activation intensity maps before and after the proposed zero-sum filters.}
\label{fig:activation_maps}
\end{figure*}

\paragraph{Feature activations before and after zero-sum filtering.}
To confirm the effect of our discrepancy transformation, we visualize feature activations immediately before and after the proposed zero-sum filter bank. Concretely, at a fixed backbone stage we take the feature tensor $F_4$ and its counterpart just after the zero-sum filters, compute an activation intensity map, and upsample it to the input size. \cref{fig:activation_maps} shows that pre zero-sum activations tend to follow more the document content, whereas after applying the zero-sum filters, the activation concentrates around local inconsistencies, producing sharper evidence near the edited region while suppressing content-driven responses. This qualitative behavior matches the role of the transformation: expose tampering boundaries as discrepancy cues that are later integrated by the multi-level fusion decoder.

\paragraph{Relation to constrained convolution residual modalities.}
Several works in image forensics use high-pass convolutions to extract a residual signal as an additional input modality. Common examples include fixed residual features such as SRM~\cite{srm} and constrained convolution layers that enforce a residual form, as in Bayar convolutions~\cite{bayar2016,bayar2018}. These residuals are typically computed from the input image and then fused with RGB features or processed in a separate branch. Our approach is fundamentally different. First, our multi-level discrepancy transformation is not an extra modality. It is applied to the backbone feature maps at multiple pyramid stages, before multi-level fusion and prediction, so it directly shapes the representation used for localization rather than adding a parallel input stream. Second, we take the absolute response and treat it as a discrepancy-strength signal, while residual branches typically keep signed residuals. Third, we do not fix the center weight as in Bayar convolutions, where the center is set to $-1$ and neighbor weights are normalized to sum to $1$~\cite{bayar2018}. We only enforce the zero-sum property and use two filter families, center-anchored mixtures and free zero-sum mixtures. In \cref{ablation} we test these alternatives and show that removing our pathway and adding a residual modality, or replacing our pathway with SRM kernels or Bayar-style constrained variants, reduces performance.
\subsection{Deep multi-scale fusion and mask prediction}

We fuse the transformed pyramid features in an FPN-style top-down pathway, progressively propagating information from fine to coarse stages. We then refine the fused representation with six ConvNeXt-V2 blocks. Finally, a lightweight segmentation head outputs the tampering mask at the input resolution. Since the transformed features operate at a relatively small width after $r(\cdot)$, these fusion and prediction modules add only minor overhead.

\begin{table*}[t!]
\centering
\small
\setlength{\tabcolsep}{4pt}
\renewcommand{\arraystretch}{0.95}
\caption{Results for models trained and evaluated following the Doc Protocol. Avg.\ C denotes the average F1 score across all cross-domain test sets.}
\label{tab:doc_protocol_results_cross}
\resizebox{\textwidth}{!}{%
\begin{tabular}{ll|cccc|c}
\toprule
\textbf{Method} & \textbf{Venue} &
\textbf{T-SROIE} & \textbf{OSTF} & \textbf{TPIC-13} & \textbf{RTM} &
\textbf{Avg.\ C} \\
\midrule
TruFor    & CVPR'23  & 0.213 & 0.046 & 0.104 & 0.034 & 0.099 \\
MVSS-Net  & ICCV'21  & 0.187 & 0.037 & 0.113 & 0.027 & 0.091 \\
Mesorch   & AAAI'25  & 0.294 & 0.139 & 0.241 & 0.041 & 0.179 \\
IML-ViT   & arXiv    & 0.427 & 0.21  & 0.256 & 0.076 & 0.242 \\
PSCC-Net  & TCSVT'22 & 0.517 & 0.441 & 0.55  & 0.126 & 0.408 \\
CAT-Net   & IJCV'22  & 0.609 & 0.178 & 0.343 & 0.063 & 0.298 \\
TIFDM     & TCE'24   & 0.058 & 0.006 & 0.013 & 0.018 & 0.024 \\
CAFTB-Net & TOMM'25  & 0.262 & 0.119 & 0.301 & 0.033 & 0.179 \\
DTD       & CVPR'23  & 0.525 & 0.124 & 0.284 & 0.058 & 0.247 \\
FFDN      & ECCV'24  & 0.533 & 0.241 & 0.357 & 0.071 & 0.301 \\
ADCD-Net  & ICCV'25  & 0.623 & 0.441 & 0.579 & 0.159 & 0.450  \\
DiffNet (Ours) & --- & \textbf{0.745} & \textbf{0.495} & \textbf{0.647} & \textbf{0.173} & \textbf{0.515} \\
\bottomrule
\end{tabular}%
}
\end{table*}

\begin{table*}[b!]
\centering
\small
\setlength{\tabcolsep}{2.6pt} 
\renewcommand{\arraystretch}{1.08}
\caption{Results for models trained and evaluated under the Syn2Real-TDoc-2.8M protocol. P, R, and F1 denote precision, recall, and F1 score, respectively. Avg.\ F1 reports the average F1 across RTM, FindItAgain, and FindIt, separately for image-level (Im) and pixel-level (Pix) predictions.}
\label{tab:syn2realtdoc-results}

\begin{adjustbox}{max width=\textwidth}
\begin{tabular}{ll|ccc|ccc|ccc|ccc|ccc|ccc|cc}
\toprule
\multirow{3}{*}{\textbf{Model}} & \multirow{3}{*}{\textbf{Venue}}
& \multicolumn{6}{c|}{\textbf{RTM}}
& \multicolumn{6}{c|}{\textbf{FindItAgain}}
& \multicolumn{6}{c|}{\textbf{FindIt}}
& \multicolumn{2}{c}{\textbf{Avg.\ F1}} \\
\cmidrule(lr){3-8}\cmidrule(lr){9-14}\cmidrule(lr){15-20}\cmidrule(lr){21-22}
& & \multicolumn{3}{c|}{\textbf{Im}} & \multicolumn{3}{c|}{\textbf{Pix}}
& \multicolumn{3}{c|}{\textbf{Im}} & \multicolumn{3}{c|}{\textbf{Pix}}
& \multicolumn{3}{c|}{\textbf{Im}} & \multicolumn{3}{c}{\textbf{Pix}}
& \textbf{Im} & \textbf{Pix} \\
\cmidrule(lr){3-5}\cmidrule(lr){6-8}
\cmidrule(lr){9-11}\cmidrule(lr){12-14}
\cmidrule(lr){15-17}\cmidrule(lr){18-20}
& & P & R & F1 & P & R & F1
& P & R & F1 & P & R & F1
& P & R & F1 & P & R & F1 & & \\
\midrule
PSCC-Net   & TCSVT'22 & 0.393 & 0.402 & 0.397 & 0.105 & 0.108 & 0.106 & 0.567 & 0.600 & 0.583 & 0.106 & 0.128 & 0.116 & 0.629 & 0.550 & 0.587 & 0.120 & 0.121 & 0.120 & 0.522 & 0.114 \\
CAT-Net    & IJCV'22  & 0.400 & 0.350 & 0.374 & 0.092 & 0.110 & 0.100 & 0.480 & 0.342 & 0.400 & 0.110 & 0.056 & 0.074 & 0.515 & 0.425 & 0.466 & 0.130 & 0.105 & 0.117 & 0.413 & 0.097 \\
DTD        & CVPR'23  & 0.366 & \textbf{0.733} & 0.489 & 0.149 & 0.114 & 0.129 & 0.500 & 0.400 & 0.444 & 0.125 & 0.124 & 0.125 & 0.623 & 0.662 & 0.642 & 0.140 & 0.144 & 0.142 & 0.525 & 0.132 \\
ASC-Former & PR'25    & 0.443 & 0.323 & 0.374 & 0.234 & 0.161 & 0.191 & 0.386 & 0.629 & 0.478 & 0.153 & 0.123 & 0.136 & 0.696 & 0.688 & 0.692 & 0.301 & 0.139 & 0.190 & 0.515 & 0.172 \\
FFDN       & ECCV'24  & 0.464 & 0.510 & 0.486 & 0.282 & 0.206 & 0.238 & \textbf{0.689} & 0.571 & 0.625 & 0.316 & 0.214 & 0.255 & 0.823 & 0.700 & 0.757 & 0.363 & 0.283 & 0.318 & 0.623 & 0.270 \\
ADCD-Net   & ICCV'25  & 0.470 & 0.517 & 0.492 & 0.249 & 0.157 & 0.193 & 0.523 & \textbf{0.657} & 0.582 & 0.136 & 0.149 & 0.142 & 0.828 & 0.662 & 0.736 & 0.191 & 0.179 & 0.185 & 0.604 & 0.174 \\
\midrule
DiffNet (Ours) & ---   & \textbf{0.506} & 0.552 & \textbf{0.528} & \textbf{0.321} & \textbf{0.212} & \textbf{0.255}
              & 0.625 & 0.628 & \textbf{0.626} & \textbf{0.409} & \textbf{0.246} & \textbf{0.307}
              & \textbf{0.833} & \textbf{0.707} & \textbf{0.765} & \textbf{0.448} & \textbf{0.287} & \textbf{0.350}
              & \textbf{0.640} & \textbf{0.304} \\
\bottomrule
\end{tabular}
\end{adjustbox}

\end{table*}
\section{Experiments}
\subsection{Implementation details}
We set the DCT embedding dimension to $d_{\text{dct}}=4$, the DCT backbone depth to $6$, the discrepancy kernel size to $K=7$, and the number of discrepancy responses per channel to $M=2$, split equally across the two filter families, i.e., one free zero-sum filter and one center-anchored zero-sum filter per channel. We set the projection dimension before $\phi$ to $256$ channels at all stages, and the decoder refinement depth to $6$.
\subsection{Evaluation Protocols}
To thoroughly evaluate our architecture and demonstrate its superiority, we follow two training and evaluation protocols and compare against prior models under the same protocol.

\paragraph{Doc Protocol.}
We follow Doc Protocol's cross-domain evaluation introduced in ForensicHub~\cite{neurips}. Under this protocol, models are trained on DocTamper~\cite{Doctamper} and evaluated on T-SROIE~\cite{wang2022b}, OSTF~\cite{OSTF}, Tampered-IC13~\cite{tamperedic13}, and RTM~\cite{rtm}.
\paragraph{Syn2Real-TDoc.}
We follow the Syn2Real-TDoc protocol proposed in~\cite{leveragingcontrastive}. Under this protocol, models are trained on TDoc-2.8M and evaluated on RTM~\cite{rtm}, FindIt~\cite{findit}, and FindItAgain~\cite{finditagain}.\paragraph{} Both pretraining datasets are synthetic, while all test datasets in the Syn2Real-TDoc protocol are human-made. In particular, RTM and FindItAgain are both high-quality and extremely challenging. On the other hand, the Doc protocol focuses on cross-domain generalization, with only RTM being human-made. We provide more details about the pretraining and testing datasets in \cref{appendixdatasets}. For both protocols, we follow their respective training and evaluation procedures. We note that the reported metrics are not computed in the same way: Syn2Real first aggregates precision and recall globally and then derives F1, whereas Doc Protocol averages F1 across images and excludes samples with no positive pixels. Additional details are provided in \cref{protoocls}.

\subsection{Main Results}

\paragraph{Results on the Doc Protocol.}
Table~\ref{tab:doc_protocol_results_cross} reports cross-domain performance under the ForensicHub Doc Protocol, comparing DiffNet against TruFor~\cite{Trufor}, MVSS-Net~\cite{MVSSNet}, Mesorch~\cite{Mesoscopic}, IML-ViT~\cite{IMLViT}, PSCC-Net~\cite{PSCC-Net}, CAT-Net~\cite{CAT-Netv2}, TIFDM~\cite{newdataandopenopen}, CAFTB-Net~\cite{CAFTBNet}, DTD~\cite{Doctamper}, FFDN~\cite{dtdv2freq}, and ADCD-Net~\cite{adcd-net}. Our method achieves the best results on all test sets and improves the cross-domain average F1 from 0.45 to 0.515. 
\begin{table}[t]
\centering
\setlength{\tabcolsep}{4.2pt}
\renewcommand{\arraystretch}{1.15}
\caption{Efficiency and localization performance for state-of-the-art models. Inference and training throughput are reported in it/s, at a resolution of $768\times768$. Inference throughput is reported with batch size 1. Doc Avg F1 is the average localization F1 on the cross-domain Doc Protocol test sets. TDoc Avg F1 is the average localization F1 on Syn2Real-TDoc, and Avg all is their mean. Note that ADCD-Net has 23.7M trainable parameters and 22M frozen parameters.}
\label{tab:throughput_params_f1}
\resizebox{\linewidth}{!}{%
\begin{tabular}{r l r r c c c}
\toprule
\textbf{Params (M)} & \textbf{Model} &
\textbf{Inf. (it/s)} & \textbf{Train (it/s)} &
\textbf{Doc Avg F1} & \textbf{TDoc Avg F1} & \textbf{Avg all} \\
\midrule
  3.7  & PSCC-Net    & 15.4 & 10.0 & 0.408 & 0.114 & 0.261 \\
114.0  & CAT-Net     & 29.9 & 13.8 & 0.298 & 0.097 & 0.198 \\
 67.3  & DTD         & 29.4 & 25.3 & 0.247 & 0.132 & 0.190 \\
 79.6  & ASC-Former  & 29.6 & 27.9 &  ---  & 0.172 &  ---  \\
 45.7 & ADCD-Net     &  5.7 &  4.4 & 0.450 & 0.174 & 0.312 \\
 140.0  & FFDN        & 33.1 & 20.8 & 0.301 & 0.270 & 0.286 \\
113.0  & DiffNet (Ours)        & \textbf{40.1} & \textbf{41.9} & \textbf{0.515} & \textbf{0.304} & \textbf{0.410} \\
\bottomrule
\end{tabular}%
}
\end{table}
\begin{table}[b]
\centering
\small
\setlength{\tabcolsep}{1pt}
\renewcommand{\arraystretch}{1.06}
\caption{Core ablations. Doc Pix F1 and TDoc Pix F1 denote the average pixel-level F1 under the cross-domain evaluation of Doc Protocol and Syn2Real-TDoc, respectively. Avg.\ Pix F1 is the equal-weight mean of the two protocol averages. All values reported are averaged across three different runs.}
\label{tab:ablation_core_pix}
\resizebox{\linewidth}{!}{
\begin{tabular}{l l c c c}
\toprule
\textbf{ID} & \textbf{Ablation} & \textbf{Doc Pix F1} & \textbf{TDoc Pix F1} & \textbf{Avg.\ Pix F1} \\
\midrule
B  & baseline (ours) & \textbf{0.515} (±0.005) & \textbf{0.303} (±0.003) & \textbf{0.409} \\
\midrule
A0  & w/o zero-sum constraint & 0.496 (±0.007) & 0.291 (±0.004) & 0.393 \\
A1  & replace our DCT backbone with FPH (comparable params/FLOPs) & 0.493 (±0.019) & 0.290 (±0.006)  & 0.391 \\
A2 & w/o fused representation refinement (compute added to the last stage instead) & 0.508 (±0.007) & 0.294 (±0.006)  & 0.401 \\
\bottomrule
\end{tabular}}
\end{table}

\paragraph{Results on Syn2Real-TDoc.}
Table~\ref{tab:syn2realtdoc-results} reports performance under the Syn2Real-TDoc protocol. Our method achieves the best overall performance, reaching 0.640 average image-level F1 and 0.304 average pixel-level F1.
Compared with the strongest prior baseline, this corresponds to gains of +0.017 and +0.034 respectively.
    The improvements are consistent across datasets and are particularly pronounced for pixel-level localization. On RTM we improve pixel-level F1 from 0.238 to 0.255, on FindItAgain from 0.255 to 0.307. Both of these datasets are human-made and extremely challenging. They contain carefully crafted manipulations, often with advanced concealment techniques designed to hide forensic traces. They also include a much higher proportion of pristine images than manipulated ones, which makes achieving high metrics difficult. Nevertheless, our method consistently improves over the previous best on both datasets.

\paragraph{Efficiency and robustness}
Table~\ref{tab:throughput_params_f1} summarizes throughput and the average localization accuracy across both protocols. Our model largely surpasses previous work on both protocols, while being substantially faster than previous work. It achieves 40.1 it/s at inference and 41.9 it/s during training, outperforming the fastest prior high-performing baseline FFDN and providing up to a $7\times$ inference speedup over ADCD-Net.

\subsection{Ablation Studies}
\label{ablation}
We perform ablation studies to quantify the contribution of each component in our design. All reported results are averaged over three runs.


\begin{table*}[t]
\centering
\small
\setlength{\tabcolsep}{12pt}
\renewcommand{\arraystretch}{1.06}
\caption{Additional ablation results. All values reported are averaged across three different runs.}
\label{tab:ablation_other_doc_pix}
\resizebox{\textwidth}{!}{%
\begin{tabular}{l l c}
\toprule
\textbf{ID} & \textbf{Ablation} & \textbf{Doc Pix F1} \\
\midrule
B & baseline (ours) & \textbf{0.515} (±0.005) \\
\midrule
\multicolumn{3}{l}{\textit{DCT embedding design}} \\
A3  & w/o table bias $t$ & 0.513 (±0.003) \\
A4  & w/o frequency-index embedding $f$ & 0.512 (±0.013) \\
A5  & w/o $\gamma$ & 0.509 (±0.014) \\
A6  & w/o $\beta$ & 0.504 (±0.017) \\
A7  & linear instead of embedding & 0.494 (±0.007) \\
\midrule
\multicolumn{3}{l}{\textit{DCT embedding dimension}} \\
A8  & $d_{\mathrm{dct}}=1$ & 0.507 (±0.008) \\
A9  & $d_{\mathrm{dct}}=2$ & 0.511 (±0.007) \\
A10 & $d_{\mathrm{dct}}=8$ & 0.515 (±0.011) \\
\midrule
\multicolumn{3}{l}{\textit{Discrepancy filters}} \\
A11 & zero-sum kernel size $K: 7 \rightarrow 9$ & 0.514 (±0.009) \\
A12 & zero-sum kernel size $K: 7 \rightarrow 5$ & 0.512 (±0.006) \\
A13 & zero-sum kernel size $K: 7 \rightarrow 3$ & 0.512 (±0.006) \\
A14 & fixed SRM filters instead of learned discrepancy filters & 0.511 (±0.005) \\
A15 & no absolute value after zero-sum filters & 0.510 (±0.008) \\
A16 & $\tanh$ instead of absolute value after zero-sum filters & 0.512 (±0.003) \\
A17 & one filter family only (best single-family choice) & 0.513 (±0.004) \\
A18 & use bayar-conv & 0.506 (±0.011) \\
A19 & discrepancy filters as a modality & 0.499 (±0.011) \\
A20 & w/o zero-sum constraint + more regularization & 0.497 (±0.006) \\
\bottomrule
\end{tabular}%
}
\end{table*}

\noindent Table~\ref{tab:ablation_core_pix} shows ablation results of our core design choices under both protocols. Removing the zero-sum constraint (A0) lowers performance, which validates the effectiveness of the discrepancy transformation. Swapping our DCT backbone for the Frequency Perception Head at matched compute (A1) leads to a similar drop, showing that our compact DCT--quantization joint embedding provides a better representation. Finally, removing the fused-representation refinement (A2) causes a smaller but consistent decrease.

\noindent Table~\ref{tab:ablation_other_doc_pix} covers several design choices of our approach. For the DCT embedding, removing the global table bias $t$ (A3) or the frequency-index embedding $f$ (A4) reduces performance, while removing the FiLM modulation terms $\gamma$ (A5) or $\beta$ (A6) causes a larger performance drop. This confirms the importance of each term of our joint embedding formula. Replacing the embedding with a linear mapping (A7) produces a large decrease, which confirms that learning an embedding is more efficient. Varying the embedding dimension (A8-9-10) shows that the used value of four provides the best efficiency.

\noindent The discrepancy-filter ablation results in Table~\ref{tab:ablation_other_doc_pix} confirm the contribution of different used values and design choices. Varying the kernel size (A11-12-13) shows that the used kernel size of seven provides the best performance and efficiency. Using fixed SRM kernels (A14) or Bayar convolution layers (A18) reduces performance compared to our proposed learned discrepancy filters. Removing the absolute-value operator (A15) or replacing it with $\tanh$ (A16) does also degrade performance, showing that taking the absolute values of the response of these zero-sum filters is beneficial. Using only one filter family (A17) also reduces accuracy, confirming that the free and center-anchored filters complement each other. Finally, treating discrepancy filters as an additional modality (A19) rather than transforming the feature pyramid or removing the zero-sum filters and replacing them with more regularization (A20) reduces performance.  
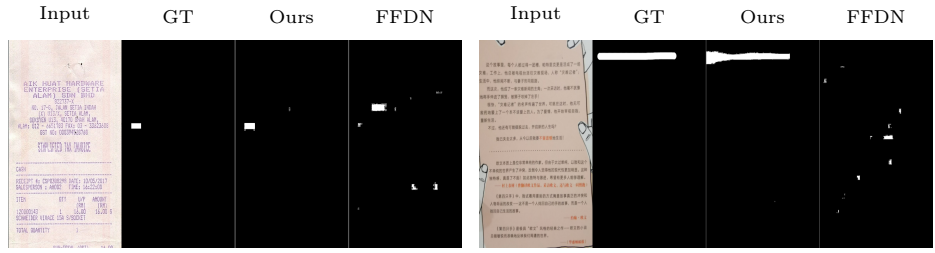
\begin{figure*}[t!]
\centering

\begin{tikzpicture}
  \node[anchor=south west, inner sep=0] (img) {\includegraphics[width=0.49\textwidth]{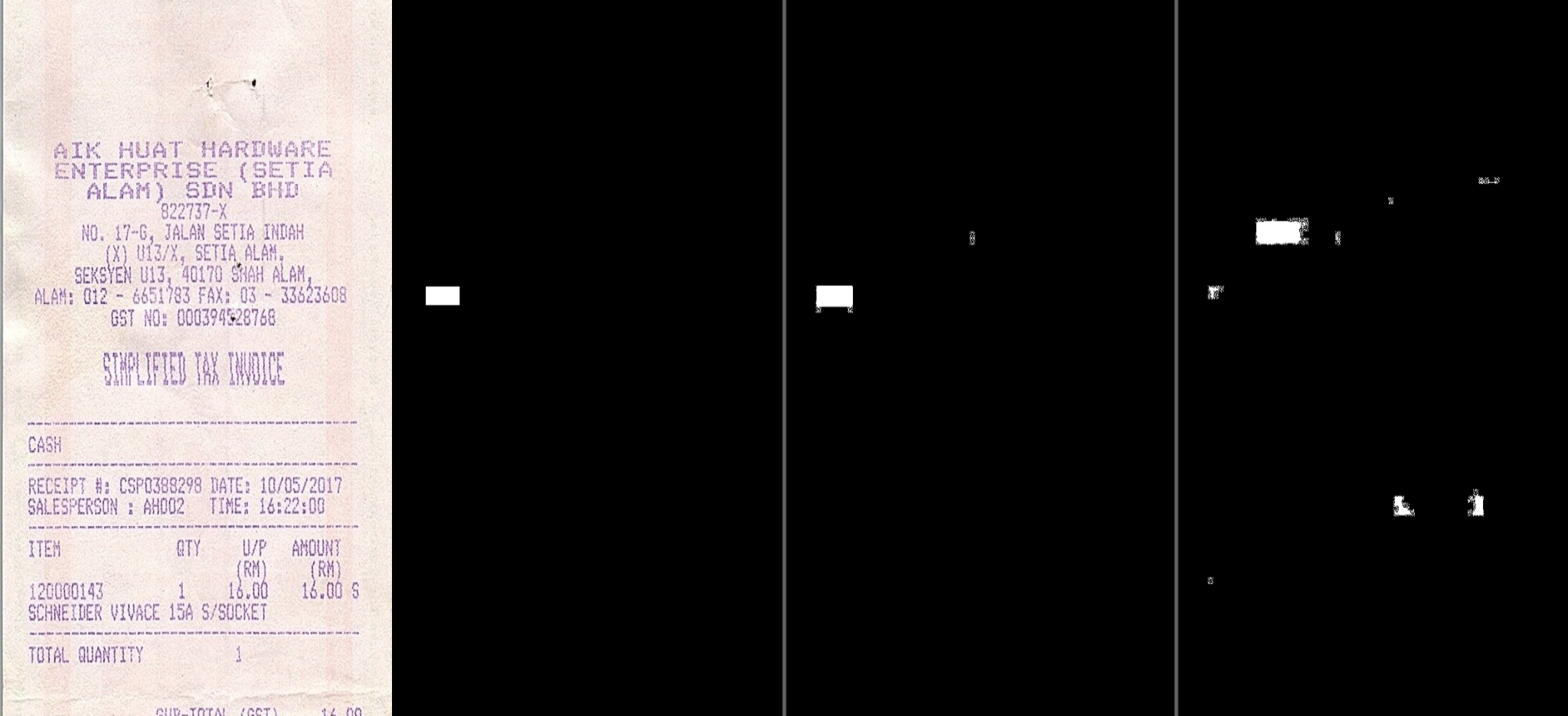}};
  \node[anchor=south, font=\scriptsize] at ($(img.north west)!0.125!(img.north east)+(0,1.2mm)$) {Input};
  \node[anchor=south, font=\scriptsize] at ($(img.north west)!0.375!(img.north east)+(0,1.2mm)$) {GT};
  \node[anchor=south, font=\scriptsize] at ($(img.north west)!0.625!(img.north east)+(0,1.2mm)$) {Ours};
  \node[anchor=south, font=\scriptsize] at ($(img.north west)!0.875!(img.north east)+(0,1.2mm)$) {FFDN};
\end{tikzpicture}\hfill
\begin{tikzpicture}
  \node[anchor=south west, inner sep=0] (img) {\includegraphics[width=0.49\textwidth]{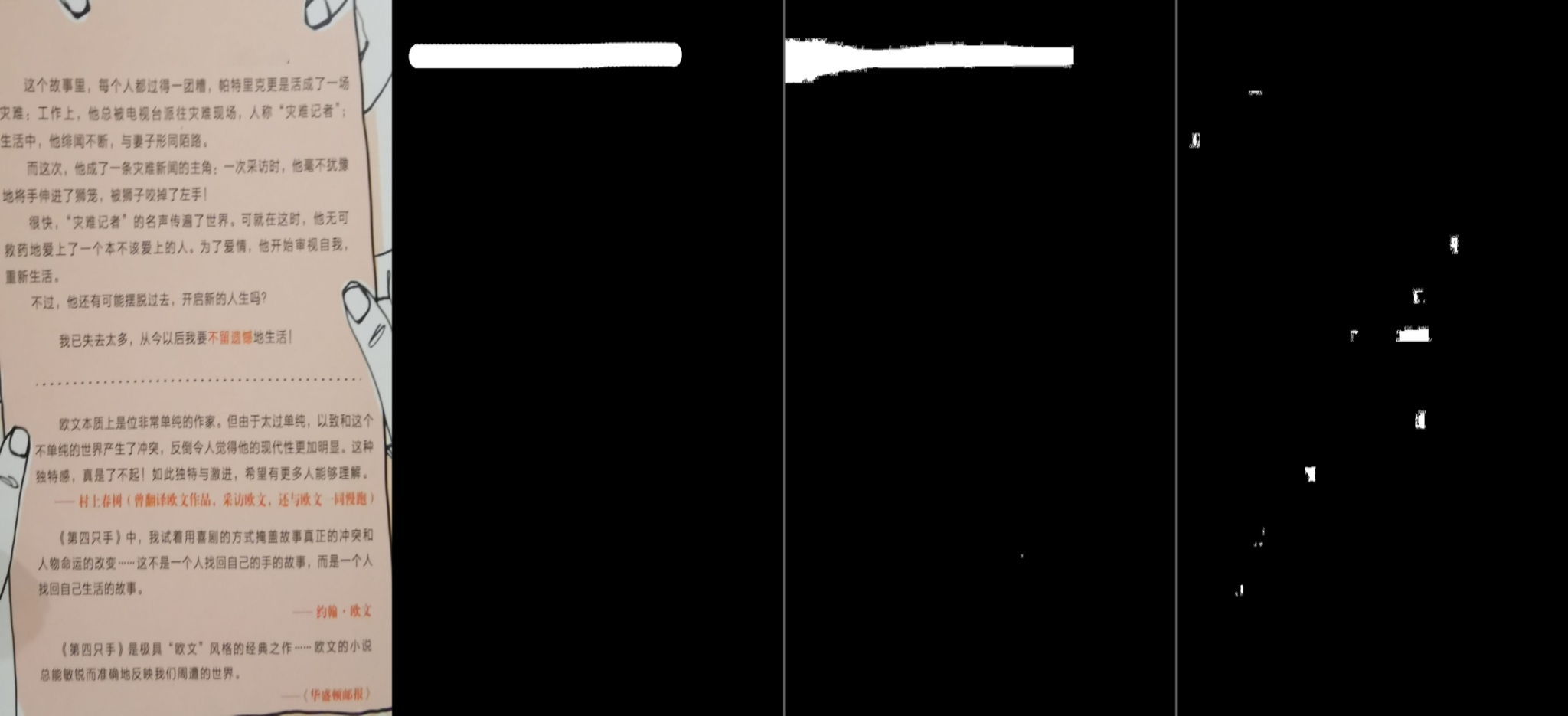}};
  \node[anchor=south, font=\scriptsize] at ($(img.north west)!0.125!(img.north east)+(0,1.2mm)$) {Input};
  \node[anchor=south, font=\scriptsize] at ($(img.north west)!0.375!(img.north east)+(0,1.2mm)$) {GT};
  \node[anchor=south, font=\scriptsize] at ($(img.north west)!0.625!(img.north east)+(0,1.2mm)$) {Ours};
  \node[anchor=south, font=\scriptsize] at ($(img.north west)!0.875!(img.north east)+(0,1.2mm)$) {FFDN};
\end{tikzpicture}

\caption{Comparison on two examples of our model and the previous state-of-the-art method, FFDN, with both models pretrained on TDoc-2.8M.}
\label{fig:qual_rtm_vs_ffdn}
\end{figure*}

\subsection{Qualitative analysis}
\label{sec:qualitative}

\paragraph{Qualitative comparison}
\cref{fig:qual_rtm_vs_ffdn} shows two representative examples comparing our model against FFDN. In these examples, our predictions tend to better concentrate on the edited area and recover a more coherent manipulated region, while suppressing scattered responses on unrelated content, which is consistent with the role of our multi-level discrepancy transformation.
\paragraph{Hard cases.}
\begin{figure*}[b!]
\centering

\begin{tikzpicture}
  \node[anchor=south west, inner sep=0] (img) {\includegraphics[width=0.49\textwidth]{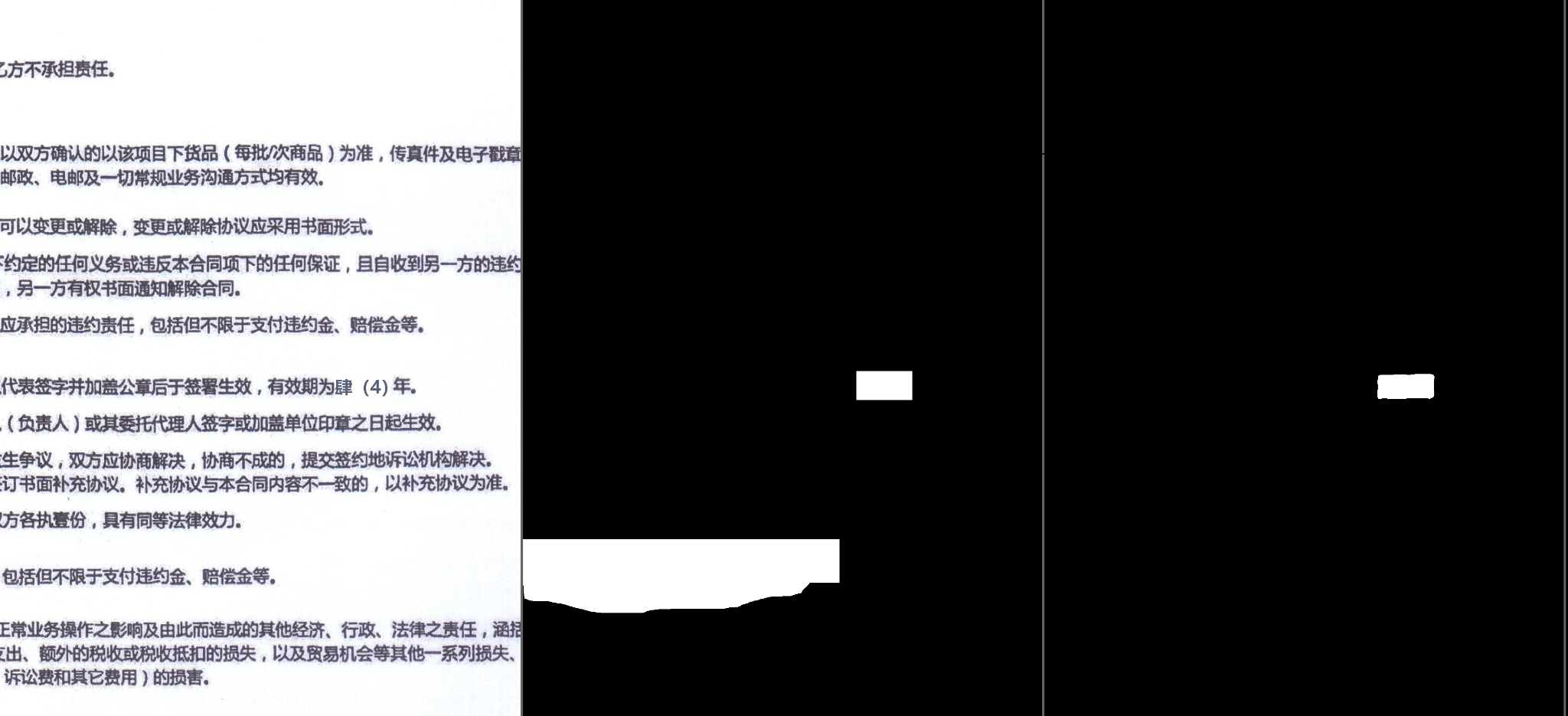}};
  \node[anchor=south, font=\scriptsize] at ($(img.north west)!0.1667!(img.north east)+(0,1.2mm)$) {Input};
  \node[anchor=south, font=\scriptsize] at ($(img.north west)!0.5000!(img.north east)+(0,1.2mm)$) {GT};
  \node[anchor=south, font=\scriptsize] at ($(img.north west)!0.8333!(img.north east)+(0,1.2mm)$) {Ours};
\end{tikzpicture}\hfill
\begin{tikzpicture}
  \node[anchor=south west, inner sep=0] (img) {\includegraphics[width=0.49\textwidth]{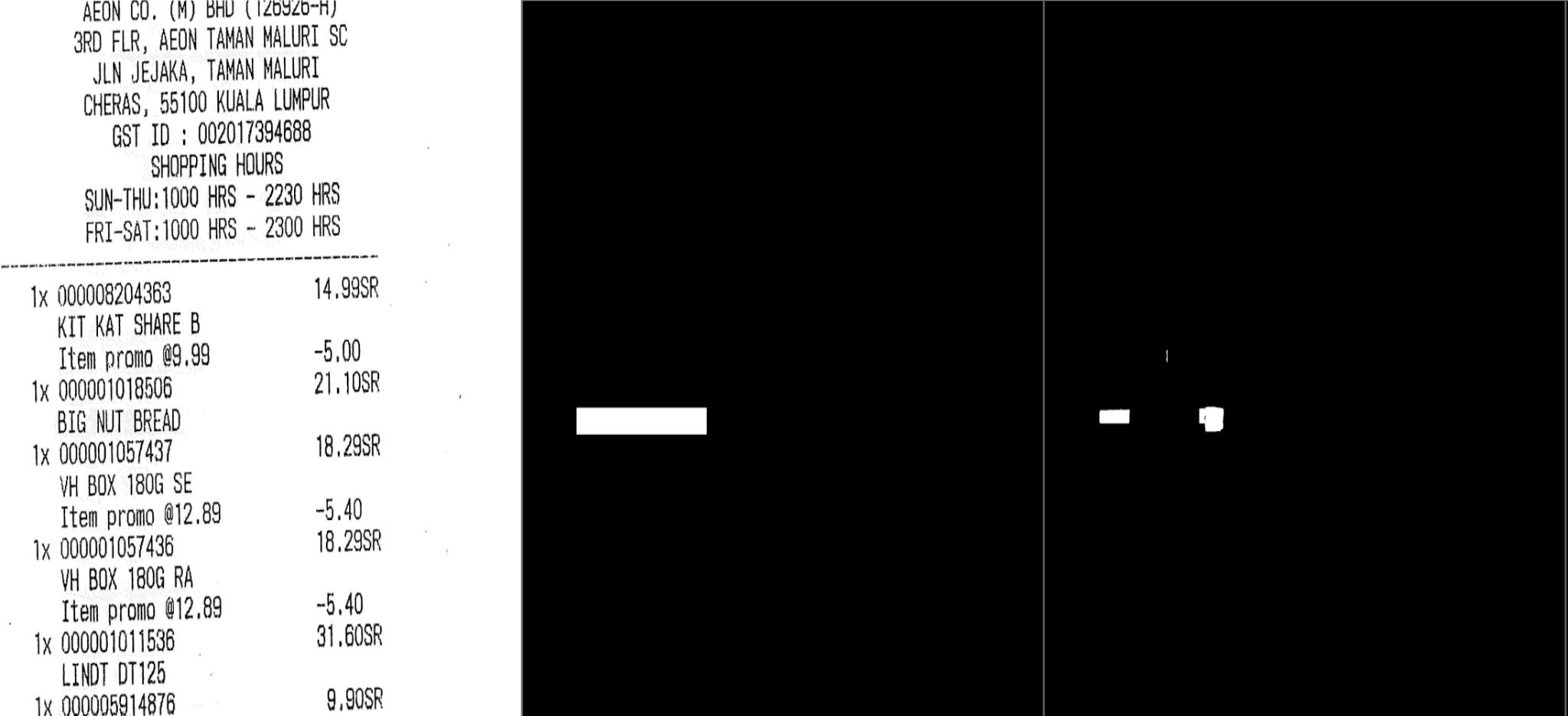}};
  \node[anchor=south, font=\scriptsize] at ($(img.north west)!0.1667!(img.north east)+(0,1.2mm)$) {Input};
  \node[anchor=south, font=\scriptsize] at ($(img.north west)!0.5000!(img.north east)+(0,1.2mm)$) {GT};
  \node[anchor=south, font=\scriptsize] at ($(img.north west)!0.8333!(img.north east)+(0,1.2mm)$) {Ours};
\end{tikzpicture}

\caption{Illustration of hard cases where our model struggles to recover the full manipulated region.}
\label{fig:qual_rtm_hard}
\end{figure*}
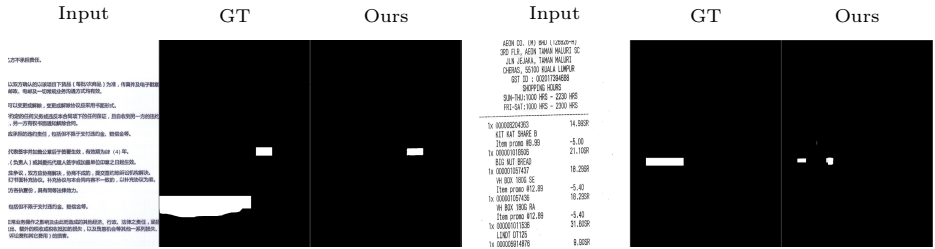
\cref{fig:qual_rtm_hard} illustrates hard samples where our model only partially localizes the forgery, yielding very low recall. These cases are challenging because the edits occur in areas with blank background, and are locally post-processed to conceal traces of tampering, leaving very few cues about the forgery location.

\paragraph{Discussion.} Our method achieves a consistent boost in performance compared to prior work. However, the localization performance on human-made tampering remains low. This is due to several factors. First, these datasets contain many challenging cases where very few tampering cues remain: edits are often performed in homogeneous background regions and combined with advanced concealment techniques, making high recall difficult. In contrast, synthetic datasets typically lack concealment strategies that adapt their type and strength to each individual case. While some works attempt to conceal tampering with fixed rules \cite{newdataandopenopen}, this can introduce shortcuts \cite{leveragingcontrastive} during training. Improving the realism and diversity of concealment in synthetic data is therefore a promising direction for future work. Second, these benchmarks include a large proportion of regions with no positive pixels, reflecting real-world deployment conditions and making it difficult to achieve high precision. Pixel F1 is also sensitive to label ambiguity and annotation shift: label mismatches between benchmarks and synthetic pretraining can penalize otherwise plausible predictions. 

\section{Conclusion}
We present an efficient RGB--DCT early-fusion architecture for human-made document tampering detection and localization. Our approach is built around two complementary ideas: (i) a multi-level discrepancy transformation applied at every backbone stage to convert features into a sign-invariant discrepancy representation, and (ii) a lightweight frequency-index-aware joint embedding for quantized DCT coefficients and quantization tables, enabling an efficient DCT branch for early fusion. Across both the Doc Protocol and the Syn2Real-TDoc protocol, the proposed design consistently improves performance, while also improving training and inference throughput.

\bibliographystyle{splncs04}
\bibliography{main}

\clearpage
\appendix

\section{Custom CUDA kernel for zero-sum discrepancy filters}
\label{app:custom-op}

As explained in \cref{sec:method}, for each input channel \(c\) we apply \(M\) depthwise zero-sum discrepancy filters over a \(K\times K\) neighborhood. During training, we implement this transformation with a fused custom CUDA operator that works directly on a compact center-tied representation, rather than first materializing dense constrained kernels and then applying a standard depthwise convolution. The operator has a forward path and a backward path, both implemented directly from the same compact representation.

\paragraph{Operator convention and padding.}
We use the cross-correlation convention. Let \(p=(p_y,p_x)\in\{0,\ldots,H-1\}\times\{0,\ldots,W-1\}\) denote an output position, and let \(\Delta=(\Delta_y,\Delta_x)\in\mathcal N\) denote a non-central offset in the \(K\times K\) neighborhood. Boundary accesses are resolved with reflection padding. Writing \(\rho_H\) and \(\rho_W\) for the reflection maps along the two spatial axes, we define the reflected extension
\[
\widetilde{F}_{b,c}(q_y,q_x)
=
F_{b,c}\!\bigl(\rho_H(q_y),\rho_W(q_x)\bigr).
\]
Hence, every access \(F_{b,c}(p+\Delta)\) in the implementation is understood as \(\widetilde{F}_{b,c}(p+\Delta)\). In the backward-input computation, contributions from reflected halo accesses are folded back onto their corresponding interior pixels.

\paragraph{Compact center-tied representation.}
Let \(\mathcal N\) denote the set of non-central offsets in the \(K\times K\) neighborhood. For each input channel \(c\) and discrepancy filter \(m\), we learn only the neighbor coefficients
\[
\{a_{c,m,\Delta}\}_{\Delta\in\mathcal N},
\]
and tie the center coefficient as
\[
k_{c,m}(\mathbf 0)=-\sum_{\Delta\in\mathcal N} a_{c,m,\Delta}.
\]
Defining
\[
s_{c,m}:=\sum_{\Delta\in\mathcal N} a_{c,m,\Delta},
\]
the response at spatial location \(p\) can be written as
\[
u_{b,c,m}(p)
=
\sum_{\Delta\in\mathcal N} a_{c,m,\Delta}\,\widetilde{F}_{b,c}(p+\Delta)
-
s_{c,m}\,F_{b,c}(p).
\]

\paragraph{CUDA organization.}
Each thread block is mapped to one \((b,c)\) pair and one spatial tile. The grid dimensions are
\[
\bigl(B\!\times\! C,\;\lceil H/T_H\rceil,\;\lceil W/T_W\rceil\bigr),
\]
where \(T_H\times T_W\) is the tile size. Within the block, thread \((t_y,t_x)\) is responsible for output location
\[
p=(p_y,p_x)=(\mathrm{tile}_y+t_y,\;\mathrm{tile}_x+t_x).
\]
The tile dimensions are selected at launch time from a small set of specialized kernels according to the spatial resolution and kernel size. Concretely, we use \(16\times16\) tiles when \(\min(H,W)\ge 64\) and \(K\le 5\), \(16\times8\) tiles when \(\min(H,W)\ge 64\) and \(K=7\), and \(8\times8\) tiles otherwise. This keeps the shared-memory footprint
\[
(T_H+2r)(T_W+2r)\cdot \mathrm{sizeof(float)} + M(K^2-1)\cdot \mathrm{sizeof(float)}
\]
within budget while maintaining sufficient block count and occupancy on late stages.

The kernel size \(K\) and the number of filters \(M\) are compile-time template parameters. The loops over \(\Delta\in\mathcal N\) and \(m\in\{1,\ldots,M\}\) are fully unrolled with \texttt{\#pragma unroll}, keeping the \(M\) accumulators register-resident. The offsets are not stored in memory: for each supported \(K\), the compiler emits the corresponding static spatial displacements directly in the unrolled code.

Before the spatial computation begins, the feature tile together with the reflected border of radius \(r=(K-1)/2\) is loaded into shared memory. The coefficients \(\{a_{c,m,\Delta}\}\) for the active channel \(c\) are also cooperatively loaded into shared memory and then reused by all threads in the block. Since the coefficients depend only on \(c\) and not on the batch index \(b\), blocks with different batch indices but the same channel also benefit from L2 cache reuse.

\paragraph{Forward path.}
At the beginning of each forward call, a lightweight preprocessing kernel computes the scalar sums
\[
s_{c,m}=\sum_{\Delta\in\mathcal N} a_{c,m,\Delta}
\]
for all \((c,m)\) and stores them in a contiguous buffer reused by the forward and backward kernels. The forward CUDA path then evaluates the operator directly from the compact coefficients. For each output location \(p\), the repeated inner loop runs only over the offsets \(\Delta\in\mathcal N\) and accumulates the neighbor contribution
\[
\sum_{\Delta\in\mathcal N} a_{c,m,\Delta}\,\widetilde{F}_{b,c}(p+\Delta).
\]
The center contribution is fused into the same computation through the single correction
\[
-\,s_{c,m}F_{b,c}(p),
\]
applied once per output.

This organization is preferable to evaluating
\[
\sum_{\Delta\in\mathcal N} a_{c,m,\Delta}\bigl(\widetilde{F}_{b,c}(p+\Delta)-F_{b,c}(p)\bigr)
\]
literally, since that would introduce one subtraction involving \(F_{b,c}(p)\) for every offset \(\Delta\). In the implemented form, the repeated inner loop remains a pure neighbor multiply-accumulate, while the shared center contribution is handled once per output.

\paragraph{Backward path.}
In backpropagation, gradients are computed directly from the same compact representation, without reconstructing the dense constrained kernels. Let
\[
g_{b,c,m}(p)=\frac{\partial\mathcal L}{\partial u_{b,c,m}(p)}
\]
be the upstream gradient. The gradient with respect to each learned neighbor coefficient is
\[
\frac{\partial \mathcal L}{\partial a_{c,m,\Delta}}
=
\sum_{b}\sum_{p}
g_{b,c,m}(p)\bigl(\widetilde{F}_{b,c}(p+\Delta)-F_{b,c}(p)\bigr),
\qquad \Delta\in\mathcal N,
\]
which we rewrite as
\[
\frac{\partial \mathcal L}{\partial a_{c,m,\Delta}}
=
\underbrace{\sum_{b}\sum_{p} g_{b,c,m}(p)\widetilde{F}_{b,c}(p+\Delta)}_{C_{c,m,\Delta}}
-
\underbrace{\sum_{b}\sum_{p} g_{b,c,m}(p)F_{b,c}(p)}_{B_{c,m}}.
\]
Since \(B_{c,m}\) does not depend on \(\Delta\), it is computed once for each \((c,m)\) and reused for all offsets.

The spatial sums \(B_{c,m}\) and \(C_{c,m,\Delta}\) are computed with a two-phase reduction. In the first phase, each thread block processes one spatial tile and accumulates tile-local partial sums using warp-level and block-level reductions; these tile-partial sums are written to a compact scratch buffer. In the second phase, a lightweight reduction kernel sums the tile-partials into the final gradients. This avoids atomic additions on the critical path and yields deterministic results for a fixed launch configuration and reduction order, up to the usual floating-point associativity effects.

The gradient with respect to the input feature map is the exact adjoint of the
reflected forward access. Since reflection padding maps several halo positions onto
the same interior pixel, all contributions that reflect onto a pixel \(q\) must be
folded back onto it:
\[
\frac{\partial \mathcal L}{\partial F_{b,c}(q)}
=
\sum_{m}
\left(
\sum_{\Delta\in\mathcal N} a_{c,m,\Delta}
\!\!\sum_{\substack{p\,:\,\rho(p+\Delta)=q}}\!\! g_{b,c,m}(p)
\;-\;
s_{c,m}\,g_{b,c,m}(q)
\right),
\]
where the inner sum ranges over all output positions \(p\) whose reflected access
\(\rho(p+\Delta)\) lands on \(q\). In the interior, where \(\rho(p+\Delta)=p+\Delta\),
this is a standard transposed correlation; the fold only modifies the outermost
radius-\(r\) ring. We realize it without atomics by first accumulating the
transposed correlation into a gradient buffer extended by the radius-\(r\) halo and
then folding each halo entry onto its reflected interior source.

\paragraph{Why the center-tied form is efficient.}
The main benefit of the center-tied form is not merely that the center coefficient is not learned independently. Rather, tying converts the zero-sum constraint into a small set of shared scalar quantities that can be reused throughout forward and backward. In the forward computation, the center contribution is represented by the single term \(-s_{c,m}F_{b,c}(p)\). In the backward-input computation, it becomes \(-s_{c,m}g_{b,c,m}(q)\). In the backward-weight computation, it becomes the shared term \(B_{c,m}\), reused across all offsets. Hence, tying turns zero-sum enforcement into a cheap scalar correction rather than a dense kernel transformation.

\paragraph{Why this is better than materializing dense kernels.}
A naive training-time implementation would first construct a dense \(K\times K\) kernel for every \((c,m)\), enforce the zero-sum constraint on that dense tensor, and only then apply depthwise correlation. This adds overhead that is not part of the actual spatial filtering: one must materialize the dense kernel tensor, reduce over its entries to enforce the constraint, modify those entries, and then read the constrained kernel again during convolution. The same issue appears in backpropagation if gradients are propagated through the dense constrained kernel. By contrast, our implementation stores only the compact neighbor coefficients, computes the scalar \(s_{c,m}\) once, and evaluates the constrained operator directly. Gradients are likewise computed directly from the compact coefficients and shared center terms, without reconstructing or differentiating through a dense constrained kernel.

\paragraph{Why this is preferable to mean subtraction.}
An alternative way to impose the zero-sum constraint is to start from an unconstrained dense kernel \(\tilde{k}_{c,m}\) and subtract its mean:
\[
k_{c,m}(\delta)=\tilde{k}_{c,m}(\delta)-\frac{1}{K^2}\sum_{\delta'}\tilde{k}_{c,m}(\delta').
\]
Although valid, this requires storing all \(K^2\) coefficients, reducing over all of them, and modifying all of them before convolution. In other words, the constraint is enforced by rewriting the whole kernel. In the center-tied form, only the \(K^2-1\) neighbor coefficients are stored, and the full correction is represented by the scalar \(s_{c,m}\). Thus, tying is more efficient precisely because it replaces a dense kernel transformation by one shared scalar correction.

\paragraph{Fusing the full discrepancy block.}
In the actual CUDA implementation, we fuse the full discrepancy block into a single operator: zero-sum filtering, absolute-value nonlinearity, and the subsequent per-channel projection back to the working width. For each input channel \(c\), the kernel computes the \(M\) zero-sum responses, keeps them in registers, applies the absolute value in place, and immediately combines them through the learned projection weights. Therefore, the expanded representation of size \(B\times C\times M\times H\times W\) is never materialized as an intermediate activation tensor. This is advantageous because, for small \(M\), the cost of the absolute value and the projection is negligible compared with the cost of storing and reloading the expanded responses from global memory. By fusing all three steps, the operator loads the feature tile once, performs all local computations while the \(M\) responses remain register-resident, and writes only the final projected output. 

\paragraph{Late stages and measured speedup.}
Avoiding dense-kernel materialization becomes especially beneficial in deeper backbone stages, where feature maps are spatially small while the discrepancy filters remain relatively large. For a \(512\times512\) input, stage 4 of a stride-32 backbone has spatial size \(16\times16\), i.e., \(256\) spatial locations. With \(M=2\) and \(K=7\), the dense filter bank for a single input channel already contains
\[
2\times 7\times 7 = 98
\]
coefficients. Hence, the amount of kernel data touched per channel is of the same order as the number of spatial locations in the feature map itself. In this regime, repeatedly constructing and transforming dense kernels is no longer negligible relative to the feature-map computation. Compared with a PyTorch-level center-tied baseline compiled with \texttt{torch.compile}, our custom CUDA operator increases end-to-end training throughput by about \(12\%\). Restricting the operator to the last two backbone stages still yields an acceleration of about \(11\%\).

\paragraph{Center-anchored variant.}
For the center-anchored variant, we use the same CUDA operator after replacing each neighbor coefficient by its absolute value,
\[
\tilde a_{c,m,\Delta}=|a_{c,m,\Delta}|.
\]
The tied center is then defined from
\[
s_{c,m}:=\sum_{\Delta\in\mathcal N}\tilde a_{c,m,\Delta},
\qquad
k_{c,m}(\mathbf 0)=-s_{c,m},
\]
so that the response becomes
\[
u_{b,c,m}(p)
=
\sum_{\Delta\in\mathcal N}\tilde a_{c,m,\Delta}\,\widetilde{F}_{b,c}(p+\Delta)
-
s_{c,m}\,F_{b,c}(p).
\]
Hence, the implementation is unchanged apart from applying the absolute value to the neighbor coefficients before computing the accumulated neighbor term and the tied center correction. Gradients are propagated through the reparameterization
\[
\frac{\partial \mathcal{L}}{\partial a_{c,m,\Delta}}
=
\operatorname{sign}(a_{c,m,\Delta})
\cdot
\frac{\partial \mathcal{L}}{\partial \tilde a_{c,m,\Delta}},
\]
with \(\operatorname{sign}(0)=0\).

\begin{algorithm}[t]
\caption{Forward pass of the center-tied zero-sum discrepancy operator}
\label{alg:zsd-forward}
\small
\begin{algorithmic}[1]
\Require Feature map \(F\in\mathbb{R}^{B\times C\times H\times W}\), neighbor coefficients \(\{a_{c,m,\Delta}\}_{\Delta\in\mathcal N}\), number of filters \(M\)
\Ensure Responses \(U\in\mathbb{R}^{B\times C\times M\times H\times W}\)
\State Precompute \(s_{c,m} \gets \sum_{\Delta\in\mathcal N} a_{c,m,\Delta}\) for all \((c,m)\)
\ForAll{blocks indexed by one \((b,c)\) pair and one spatial tile, in parallel}
    \State Load the tile of \(F_{b,c}\) together with its reflected border of radius \(r=(K-1)/2\) into shared memory
    \State Cooperatively load \(\{a_{c,m,\Delta}\}_{m,\Delta}\) for the active channel \(c\) into shared memory
    \ForAll{output locations \(p\) assigned to the threads of the block, in parallel}
        \For{\(m=1,\dots,M\)}
            \State \(\mathrm{acc}_m \gets 0\)
        \EndFor
        \ForAll{\(\Delta\in\mathcal N\)}
            \State \(v_\Delta \gets \widetilde{F}_{b,c}(p+\Delta)\)
            \For{\(m=1,\dots,M\)}
                \State \(\mathrm{acc}_m \gets \mathrm{acc}_m + a_{c,m,\Delta}\,v_\Delta\)
            \EndFor
        \EndFor
        \State \(v_0 \gets F_{b,c}(p)\)
        \For{\(m=1,\dots,M\)}
            \State \(U_{b,c,m}(p) \gets \mathrm{acc}_m - s_{c,m}\,v_0\)
        \EndFor
    \EndFor
\EndFor
\end{algorithmic}
\end{algorithm}

\begin{algorithm}[p]
\caption{Backward pass of the center-tied zero-sum discrepancy operator}
\label{alg:zsd-backward}
\begin{adjustbox}{max width=\linewidth,max height=.95\textheight,keepaspectratio}
\begin{minipage}{\linewidth}
\fontsize{7pt}{9pt}\selectfont
\begin{algorithmic}[1]
\Require Upstream gradients \(G=\partial\mathcal L/\partial U\), feature map \(F\), neighbor coefficients \(\{a_{c,m,\Delta}\}\), precomputed sums \(s_{c,m}\)
\Ensure Gradients \(\partial\mathcal L/\partial a_{c,m,\Delta}\) and \(\partial\mathcal L/\partial F_{b,c}\)
\State \(\Omega \gets \{0,\ldots,H-1\}\times\{0,\ldots,W-1\}\) \Comment{valid spatial domain}
\State \textbf{Phase 1 (tile-local partial sums):}
\ForAll{blocks indexed by one \((b,c)\) pair and one spatial tile, in parallel}
    \State Load the corresponding tile of \(F_{b,c}\) and \(G_{b,c,m}\) with reflected border into shared memory
    \ForAll{filters \(m=1,\dots,M\)}
        \State Accumulate tile-local
        \[
        B^{\mathrm{tile}}_{b,c,m} \gets \sum_{p\in\mathrm{tile}} G_{b,c,m}(p)\,F_{b,c}(p)
        \]
        \ForAll{\(\Delta\in\mathcal N\)}
            \State Accumulate tile-local
            \[
            C^{\mathrm{tile}}_{b,c,m,\Delta}
            \gets
            \sum_{p\in\mathrm{tile}} G_{b,c,m}(p)\,\widetilde{F}_{b,c}(p+\Delta)
            \]
        \EndFor
    \EndFor
    \State Write all tile-local partial sums to a scratch buffer
\EndFor
\State \textbf{Phase 2 (global reduction):}
\ForAll{channel-filter pairs \((c,m)\) in parallel}
    \State Reduce scratch-buffer partial sums into
    \[
    B_{c,m}=\sum_{b}\sum_{p} G_{b,c,m}(p)\,F_{b,c}(p)
    \]
    and
    \[
    C_{c,m,\Delta}=\sum_{b}\sum_{p} G_{b,c,m}(p)\,\widetilde{F}_{b,c}(p+\Delta)
    \]
    \ForAll{\(\Delta\in\mathcal N\)}
        \State Set
        \[
        \frac{\partial\mathcal L}{\partial a_{c,m,\Delta}}
        \gets
        C_{c,m,\Delta}-B_{c,m}
        \]
    \EndFor
\EndFor
\State \textbf{Backward-input:}
\ForAll{blocks indexed by one \((b,c)\) pair and one spatial tile, in parallel}
    \State Load the corresponding tile of \(G_{b,c,m}\) into shared memory
    \ForAll{input positions \(q\in\Omega\) assigned to the threads of the block, in parallel}
        \State \(\displaystyle
        \frac{\partial\mathcal L}{\partial F_{b,c}(q)}
        \gets
        \sum_{m}\Bigl(
        \sum_{\Delta\in\mathcal N} a_{c,m,\Delta}
        \!\!\!\sum_{\substack{p\,:\,\rho(p+\Delta)=q}}\!\!\!
        G_{b,c,m}(p)
        \;-\;
        s_{c,m}\,G_{b,c,m}(q)
        \Bigr)\)
    \EndFor
\EndFor
\end{algorithmic}
\end{minipage}
\end{adjustbox}
\end{algorithm}

\section{Details about the training and testing datasets}
\label{appendixdatasets}

We evaluate our method under two protocols, namely the Doc Protocol introduced in ForensicHub~\cite{neurips} and the Syn2Real-TDoc protocol introduced in~\cite{leveragingcontrastive}.

\paragraph{Datasets used in the Doc Protocol.}
For the Doc Protocol, training is performed on DocTamper~\cite{Doctamper}. DocTamper is a \emph{synthetic} document tampering dataset built using a rule-based pipeline from document images. It contains \(120{,}000\) tampered training images.

Evaluation in the Doc Protocol is then carried out on four external document manipulation datasets, namely T-SROIE, OSTF, Tampered-IC13, and RTM~\cite{neurips}. Following ForensicHub~\cite{neurips}, these datasets are converted to the same sliced format as DocTamper. We use the files provided by the authors of the protocol paper to ensure that the same evaluation is conducted.

T-SROIE~\cite{wang2022b} is a \emph{synthetic} receipt tampering dataset derived from SROIE, where the forged images are generated through text editing based on SRNet. Its sliced \emph{test split} contains \(1{,}579\) fake patches~\cite{neurips}. OSTF~\cite{OSTF} is a \emph{synthetic} text tampering benchmark in which forged images are generated by eight generative AI text editing models; its sliced \emph{test split} contains \(3{,}046\) fake patches~\cite{neurips}. Tampered-IC13~\cite{tamperedic13} is a \emph{synthetic} scene-text tampering dataset obtained by digitally altering text regions in ICDAR2013 images; its sliced \emph{test split} contains \(589\) fake patches~\cite{neurips}. RTM is a real-world text manipulation benchmark in which the manipulated images are manually edited by professional editors~\cite{rtm}. For the Doc Protocol, RTM is evaluated in the sliced format defined by ForensicHub. After applying the DocTamper-style cropping procedure, the RTM test split contains \(3{,}444\) fake patches~\cite{neurips}.

\paragraph{Datasets used in Syn2Real-TDoc.}
For Syn2Real-TDoc, we follow exactly the protocol introduced in~\cite{leveragingcontrastive}. The synthetic pretraining corpus is TDoc-2.8M, i.e., a \emph{synthetic} document tampering dataset of approximately \(2.8\) million generated tampered images~\cite{leveragingcontrastive}.

Evaluation is then performed on RTM, FindIt, and FindItAgain. RTM~\cite{rtm} is the same \emph{real-world} text manipulation benchmark described above. For the Syn2Real-TDoc protocol, its official \emph{test split} of \(3{,}197\) images is used. FindIt is a \emph{human-made} receipt forgery dataset built from scanned receipts and OCR outputs, where the forged receipts were manually created during forgery workshops using standard editing software~\cite{findit}. Its \emph{test split} contains \(549\) images. FindItAgain is also a \emph{human-made} receipt forgery benchmark. It contains \(988\) scanned receipt images derived from SROIE, among which \(163\) were manually modified to simulate realistic fraud~\cite{finditagain}. Its \emph{test split} contains \(218\) images.

As stated above, for each protocol we use exactly the same training and testing datasets, processed in exactly the same manner.
\section{Details about the training and evaluation protocols}
\label{protoocls}

\paragraph{Doc Protocol.}
Under the Doc Protocol, models are trained and evaluated on sliced \(512\times512\) patches. Following the protocol setting in~\cite{neurips}, we use a resolution of \(512\times512\) for both training and evaluation. This is the resolution used to train document tampering localization models for this protocol. Under the Doc Protocol, we follow the common benchmark recipe of ForensicHub whenever it applies. ForensicHub further specifies that the number of training iterations and the batch size depend on the model~\cite{neurips}. For settings not covered by this common recipe, we use those of the Syn2Real-TDoc Protocol.

In the Doc Protocol, pixel-level Binary F1 is computed exactly as defined by the benchmark. Specifically, they first compute Binary F1 separately for each cropped image patch, and then average these per-patch scores over the subset of patches whose ground-truth mask contains at least one forged pixel. In our paper, we follow exactly the same evaluation rule.

We note that as training and inference use the same resolution, there is no need for an inference strategy here and the crop is simply fed to the model at its native resolution, following the official protocol.

\paragraph{Syn2Real-TDoc.}
Under Syn2Real-TDoc, we follow exactly the protocol of~\cite{leveragingcontrastive}. A resolution of \(1024\times1024\) is used, exactly as in the protocol. For this protocol, unified training settings are used, including the same number of epochs, batch size, data augmentation pipeline, seed, and optimization schedule. We use exactly the same settings as those used for the models evaluated under this protocol. Training is performed for 2 epochs with a batch size of 64. The Focal Loss with \(\gamma=2\) is used for both the segmentation head and the document-level classification head. The maximum learning rate is set to \(1\times10^{-4}\), and a cosine annealing schedule is used. We also use the same data augmentation pipeline as defined in Appendix~H of~\cite{leveragingcontrastive}. In our paper, we follow exactly this protocol.

Following the Syn2Real-TDoc protocol, we use a sliding window inference strategy. For each image, inference is performed in a sliding-window manner, from which the global prediction map is computed. For the image-level prediction, the maximum over the logits of all crops is taken.

In Syn2Real-TDoc, global TP, TN, FP, and FN are computed first; precision, recall, and F1 are then derived from them. We follow the same metric for this protocol.
\end{document}